\documentclass[11pt,a4paper]{article}
\usepackage[hyperref]{acl2020}
\usepackage{times}
\usepackage{latexsym}
\usepackage{graphicx}  
\usepackage{todonotes}
\usepackage{subcaption}
\usepackage{comment}

\usepackage{microtype}

\aclfinalcopy 

\title{Let Me Choose: From Verbal Context to Font Selection}

\author{Amirreza Shirani$^{\dagger}$, 
Franck Dernoncourt$^{\ddagger}$,
Jose Echevarria$^{\ddagger}$, \\
\textbf{Paul Asente$^{\ddagger}$,
Nedim Lipka$^{\ddagger}$ 
and Thamar Solorio$^{\dagger}$} \\
 	$^{\dagger}$University of Houston \qquad
 	$^{\ddagger}$Adobe Research \qquad \\
 	$^{\dagger}$\small{\texttt{\{ashirani,tsolorio\}@uh.edu}} \\
 	$^{\ddagger}$\small{\texttt{\{franck.dernoncourt,asente,lipka,echevarr\}@adobe.com}} \\
 	}

\date{}

\begin{document}
\maketitle

\begin{abstract}
In this paper, we aim to learn associations between visual attributes of fonts and the verbal context of the texts they are typically applied to.
Compared to related work leveraging the surrounding visual context, we choose to focus only on the input text as this can enable new applications for which the text is the only visual element in the document. 
We introduce a new dataset, containing examples of different topics in social media posts and ads, labeled through crowd-sourcing. 
Due to the subjective nature of the task, multiple fonts might be perceived as acceptable for an input text, which makes this problem challenging.
To this end, we investigate different end-to-end models to learn label distributions on crowd-sourced data and capture inter-subjectivity across all annotations. 
\end{abstract}
\section{Introduction}
In visual designs, textual information requires the use of fonts with different properties. Whether it is books, magazines, flyers, ads or social media posts,
different typefaces are commonly used to express non-verbal information and add more dimensions to the text. 
%
An appropriate font usually embodies information about character, context and usage of the design \cite{doyle2006dressed}. This motivates us to explore font associations with regular users in a crowd-sourced setting. In other words, we investigate how users relate fonts to different characteristics of the input text. 

Current font selection interfaces such as \newcite{o2014exploratory} and commercial online services (e.g., MyFonts\footnote{www.myfonts.com} and Typekit\footnote{https://fonts.adobe.com/}) assist users in selecting fonts by taking into account font similarity. However, they do not consider the verbal context of the input text. 
Having a better understanding of the input text, users can benefit from a font recommendation system during authoring, saving time and avoiding tedious exploration of long lists of fonts. 

Most graphic designers agree that there is no strict or universally-accepted rule for choosing fonts. Different social and personal factors can be involved in typeface selection, which makes this process subjective. However, there seems to be enough agreement among human opinions to build reasonably effective models of font properties \cite{o2014exploratory,shinahara2019serif}. 
Several empirical studies have directly explored the relationship between fonts and texts \cite{shinahara2019serif, henderson2004impression, mackiewicz2007audience}. For example, \newcite{brumberger2003awareness} indicates that readers have strong opinions about the appropriateness of particular typefaces for particular text passages, and they can differentiate typeface/text mismatches.


In this study, we aim to model for the first time the associations between visual font attributes and textual context, with the final goal of better font recommendation during text composition.
Our main contributions are: 
1) We propose and formulate a new task: ``font recommendation from written text." 
2) We introduce a new dataset, \textit{Short Text Font Dataset}, containing a variety of text examples annotated with ten different representative fonts. 
3) We compare different end-to-end models that exploit contextual and emotional representations of the input text to recommend fonts. These models are able to capture inter-subjectivity among all annotations by learning label distributions during the training phase. 
We show that emotional representations can be successfully used to capture the underlying characteristics of sentences to suggest proper fonts. 
\section{Related Work}\label{sec:related:emotion}
Font-related studies have been extensively explored in graphic design literature. 
\newcite{shinahara2019serif} performed an empirical study on collections of book titles and online ads, showcasing trends relating typographic design and genre.
%
%
%
%
%
%
Several previous studies have attempted to associate personality traits and fonts
\cite{o2014exploratory,brumberger2003rhetoric,juni2008emotional,mackiewicz2005people,amare2012seeing}. They support the idea of typefaces consistently perceived to have particular personas, emotions, or tones. More recently, FontLex \cite{kulahcioglu2018fontlex} was the first to find the association between fonts and words
by utilizing font-emotion and word-emotion relationships.
Instead of focusing on independent words, our proposed model suggests fonts by considering the broader context of the whole text. 
%
%
%
\paragraph{Task Subjectivity}
In some tasks, aggregated annotations always correspond to the correct answer \cite{brew2010using}.
Therefore, to fully utilize the crowd's knowledge, different approaches have been proposed to aggregate labels, from simply applying majority voting to more sophisticated strategies to assess annotators' reliability \cite{yang2018leveraging,srinivasan2019crowdsourcing,rodrigues2014sequence}. All of these methods rely on the assumption that only one answer is correct and should be considered as ground truth \cite{nguyen2016probabilistic}.
Whereas in tasks like ours, sentiment analysis~\cite{brew2010using} or
facial expression~\cite{barsoum2016training}, the answer is likely to be more subjective due to its non-deterministic nature \cite{urkullu2019evaluation}. 
We follow previous studies that successfully employed label distribution learning to handle ambiguity in the annotations \cite{geng2013facial,shirani2019learning,yang2015deep}.
\section{Font Dataset}
\begin{figure}[h!]
\centering
\includegraphics[width=0.9\linewidth]{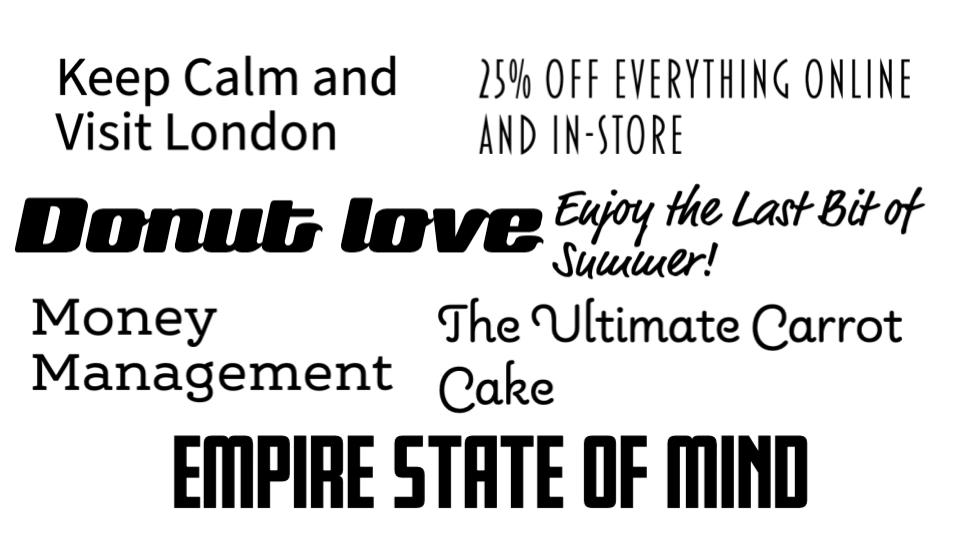}
\caption{Examples from our collected dataset visualized through fonts
with the highest annotation agreements.}
\label{fig:fonts}
\end{figure}

The proposed dataset includes 1,309 short text instances from Adobe Spark\footnote{https://spark.adobe.com.}. The dataset is a collection of publicly available sample texts created by different designers. It covers a variety of topics found in posters, flyers, motivational quotes and advertisements.\footnote{The dataset along with the annotations can be found online: \url{https://github.com/RiTUAL-UH/Font-prediction-dataset}}
\paragraph{Choice of Fonts}
A vast number of fonts and typefaces are used in contemporary printed literature. To narrow down the task, we had a font expert select a set of 10 display fonts that cover a wide range of trending styles.
These fonts display enough differentiation in visual attributes and typical use cases to cover the topics in our text samples.
%
Figure~\ref{fig:fonts} shows several examples from the dataset, each rendered with the most congruent font (font with the highest agreement).
\begin{figure*}[t]
\centering
\begin{subfigure}{0.46\textwidth}
\includegraphics[width=1\linewidth]{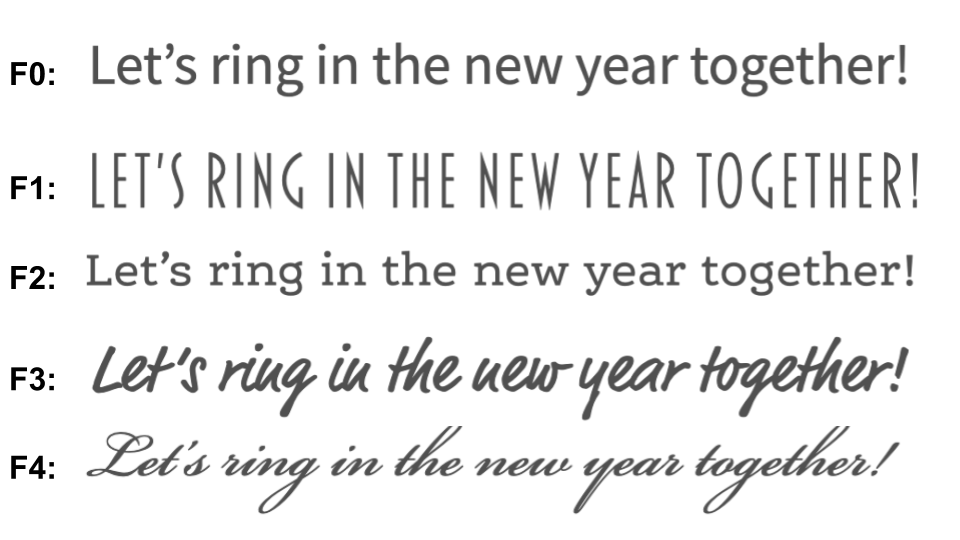} 
\end{subfigure}
\begin{subfigure}{0.46\textwidth}
\includegraphics[width=1\linewidth]{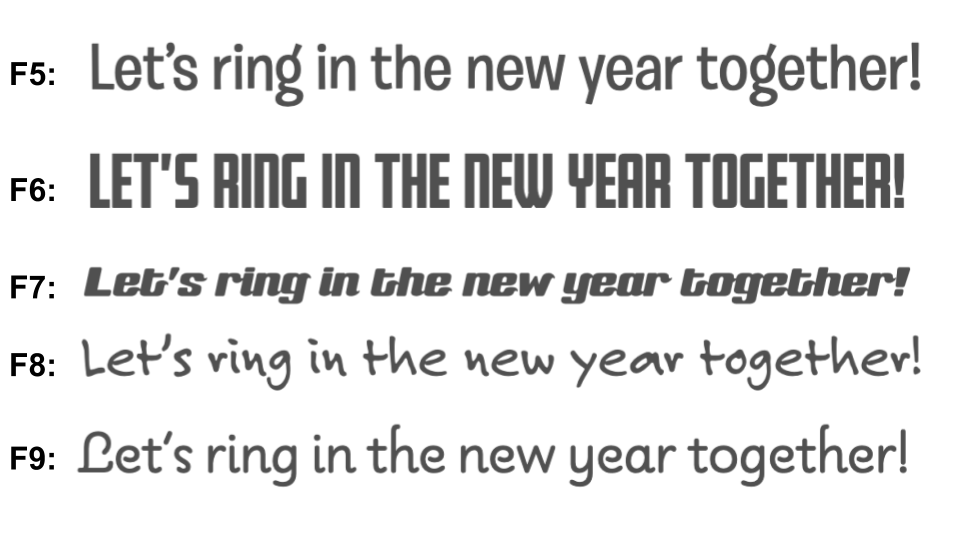}
\end{subfigure}
\caption{A text sample from the dataset rendered using the available 10 fonts for labelling. F0) Source Sans Pro, F1) Blakely, F2) FF Ernestine Pro, F3) FF Market Web, F4) Bickham Script Pro 3, F5) Burbank Big, F6) Fresno, F7) Sneakers Script Narrow, F8) Felt Tip Roman, F9) Pauline}
\label{fig:rendered}
\end{figure*}

\paragraph{Annotation Process}
In an MTurk experiment, we asked nine annotators to label each sample text by selecting their top three fonts (Figure \ref{fig:rendered}). Workers were asked to choose suitable fonts after reading the sentence. We included carefully-designed quality questions in 10 percent of the hits to monitor the quality of our labeling. 
We also needed to ensure workers selected fonts based on the comprehension of the text rather than just personal preference. Therefore, we removed the annotations of workers who selected the same font more than 90 percent of the time, resulting in six to eight annotations per instance (we removed instances with fewer than six annotations). 

\begin{figure}[htbp]
\centering
  \includegraphics[width=0.48\textwidth]{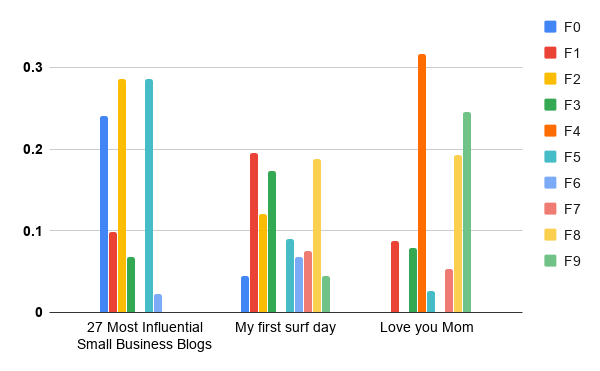}
  \caption{Label distributions for three examples}
  \label{fig:example}
\end{figure}

\begin{figure}[t]
\centering
  \includegraphics[width=0.48\textwidth]{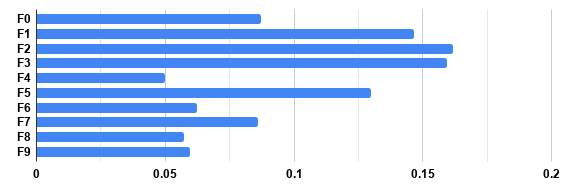}
  \caption{Average label distribution of the entire corpus}
  \label{fig:dist}
\end{figure}

As we mentioned earlier, we asked annotators to rank their top three font choices for each text in our dataset. We decided to treat the first, second, and third choices differently as they represent the workers' priorities. Therefore, we give the highest weight to the first choices (1.0) and lower weights (0.6) and (0.3) to the second and third choices, respectively. Figure~\ref{fig:example} shows three examples with label distributions over 10 fonts. 
By comparing the label distributions of these examples, we can observe that `formal' fonts like F0, F2, and F5 are often selected in business contexts (left). `modern/display' fonts like F1, F3, and F8 are favored in more casual settings (center), and `script' fonts like F4, F8, and F9 are preferred for more emotional contexts (right).
%
We observe that some fonts are more popular than others. 
Figure \ref{fig:dist} shows the average label distribution over all instances. F3, F2, and F1 are the most popular, while F4, F8, and F9 are the least popular among all 10 fonts.
\paragraph{Statistics} The dataset contains 8,211 tokens. The mean and standard deviation number of tokens per instance is 6.27 and 4.65, ranging from 1 to 27 tokens. 
We obtained a Fleiss’ kappa agreement~\cite{fleiss1971measuring} of 0.348 by taking into account all three choices. This value is reasonable for a task such as this since previous subjective tasks have also reported low inter-rater agreement scores \cite{salminen2018inter,alonso2014crowdsourcing}. 
We split up the data randomly into training (70\%), development (10\%) and test (20\%) sets for further experimentation and evaluation. 
%
\begin{figure*}[ht]
\centering
  \includegraphics[width=1\textwidth]{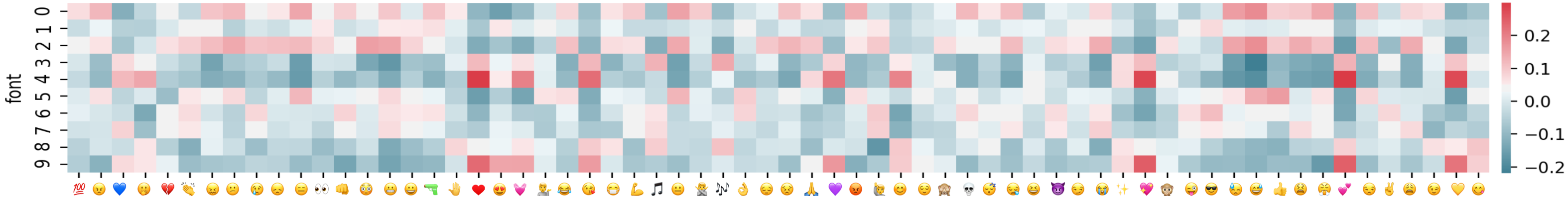}
  \caption{Font-Emoji Pearson Correlation Coefficient Heatmap}
  \label{fig:heatmap}
\end{figure*}
\section{Methodology}
\paragraph{Task Definition} Given a piece of text $X$, we want to determine which font(s) $y=\{y_0, ...y_{9}\}$ are more appropriate or congruent with the properties of the input text.
We formulate this problem as a ranking problem where the model assigns each font a real value $d^x_y$, representing the degree to which $y$ describes $X$.
In other words, $d^x_y$ represents the degree of congruency of font $y$ with input $X$. 
The values for all the labels are summed up to 1 to fully describe the instance \cite{geng2016label}. 
\subsection{Model}
We explore transfer learning from pre-trained models to improve the performance
of our task. We investigate four different deep learning-based architectures to learn font distributions of examples in our dataset. Inspired by previous works, which supported the relationship between font and emotion (Section \ref{sec:related:emotion}), we compare the effectiveness of emotional embeddings in our models to contextual embeddings like BERT.\footnote{The implementation is available online: \url{https://github.com/RiTUAL-UH/Font_LDL_2020}}
%
\paragraph{GloVe-BiLSTM Model} In this model, we use GloVe embeddings \cite{pennington2014glove} as input and a BiLSTM layer to encode word sequence information in forward and backward directions. Subsequently, we pass the encoded-words to two dense layers for prediction.
\paragraph{NRC Model} 
Similar to the GloVe-BiLSTM Model, this model is LSTM-based. The difference is that instead of GloVe embeddings, we use the emotional representations of words from NRC Emotion~\cite{mohammad2013crowdsourcing}, Intensity~\cite{LREC18-AIL} and Valence, Arousal, and Dominance (VAD)~\cite{vad-acl2018} lexicons as input to the model. To efficiently look up the emotion value of words, we search for the stemmed and synonym versions of out-of-vocabulary words.
\paragraph{BERT Model} 
We use pre-trained BERT sequence classification model~\cite{devlin2018bert} to obtain contextual embeddings as features.
Then the output is fed to two dense layers yielding the class predictions.
We implement our model based on the Hugging Face's BERT implementation \cite{Wolf2019HuggingFacesTS}.
\paragraph{Emoji Model} In this model, we use the DeepMoji pre-trained model \cite{felbo2017} to generate emoji vectors by encoding the text into 2304-dimensional feature vectors. We treat these features as embedding and pass them to the model with two dense layers.
Deepmoji\footnote{Our implementation is based on the Hugging Face Torchmoji implementation,
\newline
https://github.com/huggingface/torchMoji} is a sentence-level model containing rich representations of emotional content which is trained on a 1,246 million tweet corpus in the emoji prediction task.
\section{Experimental Settings and Results}
\subsection{Training Details}
The Kullback-Leibler Divergence (KL-DIV)~\cite{kullback1951information} is used as the loss function to train the models. KL-DIV measures how the predicted probability distribution is different from the ground truth probability distribution.
To train all the models, we use Adam optimizer \cite{kingma2014adam} to optimize the model parameters.
We run all models over four runs with different random seeds and report the averaged score to ensure stability. The reported test results correspond to models with the best accuracy on the validation set. 
\subsection{Evaluation Settings}
We evaluate the performance by using two different evaluation metrics for this new task.
\paragraph{Font Recall (FR)} 
Less popular fonts could be underrepresented by the models. Therefore we need an evaluation metric that measures the performance of models in learning individual labels.
Since we are dealing with an unbalanced dataset, motivated by evaluation methodology used in previous recommendation systems like \newcite{kar2018folksonomication, carneiro2007supervised}, we compute Font Recall, i.e. the average recall per font, to measure the performance of the models in learning individual labels.

\small\[\mbox{FR} := \frac{\sum^{|F|}_{i=1} |R_i|}{|F|}\]
\normalsize
Where $|F|$ represents the number of labels and $R_i$ is the recall for the $i^{th}$ font.
\paragraph{F-score} For each instance $X$ from the test set, we select the top $k = \{1, 3$ and $5\}$ fonts with the highest probabilities from both ground truth and prediction distributions. Then we compute weighted averaged F1-score for each $k$. 

Note that there are many cases where two or more fonts have the exact same probability. In this case, if the model predicts either one of the labels, we consider it as a correct answer in both metrics. 
%
\subsection{Results}
\begin{table}[bh]
\centering
\resizebox{0.48\textwidth}{!}{%
\begin{tabular}{l|l|l|l|l|l}
\hline
Model/Evals & FR Top3 & FR Top5 & F-Top1 & F-Top3 & F-Top5 \\ \hline
Majority Baseline & 30.00 & 50.00 & 12.44 & 43.72 & 62.24 \\ \hline
NRC Model & 30.78 & 51.60 & 23.10 & 47.27 & 66.16 \\ \hline
GloVe Model & 32.71 & 53.74 & 25.95 & 51.29 & 68.29 \\ \hline
Emoji Model & \textbf{33.17} & \textbf{54.06} & \textbf{26.00} & \textbf{51.43} & \textbf{68.53} \\ \hline
BERT Model & \textbf{33.54} & \textbf{56.00} & \textbf{26.97} & \textbf{51.91} & \textbf{69.38} \\ \hline
\end{tabular}%
}
\caption{Experimental results for all five models. FR represents Font Recall and F represents F-1 score.
The results in bold are statistically significant compared to the Majority Baseline.}
\label{tab:res}
\end{table}
Table~\ref{tab:res} compares different models in terms of five evaluation settings. The first two columns of the results show FR for the top 3 and 5 fonts. The other three columns show F-score for the top 1, 3 and 5 fonts. 
Comparing to the Majority Baseline, the results from the Emoji and BERT models are statistically significant under paired t-test with 95\% confidence interval.
Although the BERT model performs slightly better than the rest, the Emoji model performs just as well, which suggests two things: (1) the font recommendation task is highly related to what emojis represent and 2) a simpler model like Emoji model can perform similarly to a complex solution like BERT.
%

We analyze the reason behind the effectiveness of the Emoji model by visualizing the Font-Emoji Pearson Correlation Coefficient Heatmap (Figure~\ref{fig:heatmap}) in the training set. Interestingly, fonts F4 and F9 with a `Script' style are highly correlated by `Heart' and `Love' emojis. Also, F3 with a `Playful' style is negatively correlated with emojis with discomfort and mild irritation expressions.
\paragraph{Data Augmentation} 
A well-established technique for automatic data augmentation is leveraging machine translation to find meaning-equivalent phrases in a single language \cite{mallinson2017paraphrasing, coulombe2018text}. 
To mitigate the highly imbalanced class distribution in our data set, we tried over- and under-sampling techniques. We selected examples with high values in underrepresented classes and translated them to four non-English languages using Google Translate\footnote{https://cloud.google.com/translate/docs/apis}. We then translated these examples back to English, resulting in 170 more examples. We also removed 50 instances with high values in the popular classes. We observed that the data augmentation process has marginal improvements (up to 1\%) in some models. We leave the exploration of more sophisticated data augmentation approaches for future work.
%
\section{Conclusion}
In this paper, we associated font with written text and tackle the problem of font recommendation from the input text. We collected more than 1,300 short written texts and annotated them with ten fonts.
We formulated this task as a ranking problem and compared different models based on emotional and contextual representations that exploit label distribution learning to predict fonts. 

The current approach covers a fixed number of fonts, but it can be extended to support a larger set of fonts.  For example, we can use font similarity techniques and enable users to pick a group of fonts, or to provide increased flexibility for the fonts available to users.
%
\section*{Acknowledgments}
This research began during an internship at Adobe Research, and was sponsored in part by Adobe Research.
We thank the reviewers for their thoughtful comments and efforts towards improving our work.
We also thank Tim Brown for his help with font set selection.
\bibliography{anthology,acl2020}

\begin{thebibliography}{38}
\expandafter\ifx\csname natexlab\endcsname\relax\def\natexlab#1{#1}\fi

\bibitem[{Alonso et~al.(2014)Alonso, Marshall, and
  Najork}]{alonso2014crowdsourcing}
Omar Alonso, Catherine Marshall, and Marc Najork. 2014.
\newblock Crowdsourcing a subjective labeling task: a human-centered framework
  to ensure reliable results.
\newblock \emph{Microsoft Res., Redmond, WA, USA, Tech. Rep. MSR-TR-2014--91}.

\bibitem[{Amare and Manning(2012)}]{amare2012seeing}
Nicole Amare and Alan Manning. 2012.
\newblock Seeing typeface personality: Emotional responses to form as tone.
\newblock In \emph{2012 IEEE International Professional Communication
  Conference}, pages 1--9. IEEE.

\bibitem[{Barsoum et~al.(2016)Barsoum, Zhang, Ferrer, and
  Zhang}]{barsoum2016training}
Emad Barsoum, Cha Zhang, Cristian~Canton Ferrer, and Zhengyou Zhang. 2016.
\newblock Training deep networks for facial expression recognition with
  crowd-sourced label distribution.
\newblock In \emph{Proceedings of the 18th ACM International Conference on
  Multimodal Interaction}, pages 279--283. ACM.

\bibitem[{Brew et~al.(2010)Brew, Greene, and Cunningham}]{brew2010using}
Anthony Brew, Derek Greene, and P{\'a}draig Cunningham. 2010.
\newblock Using crowdsourcing and active learning to track sentiment in online
  media.
\newblock In \emph{ECAI}, pages 145--150.

\bibitem[{Brumberger(2003{\natexlab{a}})}]{brumberger2003awareness}
Eva~R Brumberger. 2003{\natexlab{a}}.
\newblock The rhetoric of typography: The awareness and impact of typeface
  appropriateness.
\newblock \emph{Technical communication}, 50(2):224--231.

\bibitem[{Brumberger(2003{\natexlab{b}})}]{brumberger2003rhetoric}
Eva~R Brumberger. 2003{\natexlab{b}}.
\newblock The rhetoric of typography: The persona of typeface and text.
\newblock \emph{Technical communication}, 50(2):206--223.

\bibitem[{Carneiro et~al.(2007)Carneiro, Chan, Moreno, and
  Vasconcelos}]{carneiro2007supervised}
Gustavo Carneiro, Antoni~B Chan, Pedro~J Moreno, and Nuno Vasconcelos. 2007.
\newblock Supervised learning of semantic classes for image annotation and
  retrieval.
\newblock \emph{IEEE transactions on pattern analysis and machine
  intelligence}, 29(3):394--410.

\bibitem[{Coulombe(2018)}]{coulombe2018text}
Claude Coulombe. 2018.
\newblock Text data augmentation made simple by leveraging nlp cloud apis.
\newblock \emph{arXiv preprint arXiv:1812.04718}.

\bibitem[{Devlin et~al.(2018)Devlin, Chang, Lee, and
  Toutanova}]{devlin2018bert}
Jacob Devlin, Ming-Wei Chang, Kenton Lee, and Kristina Toutanova. 2018.
\newblock Bert: Pre-training of deep bidirectional transformers for language
  understanding.
\newblock \emph{arXiv preprint arXiv:1810.04805}.

\bibitem[{Doyle and Bottomley(2006)}]{doyle2006dressed}
John~R Doyle and Paul~A Bottomley. 2006.
\newblock Dressed for the occasion: Font-product congruity in the perception of
  logotype.
\newblock \emph{Journal of consumer psychology}, 16(2):112--123.

\bibitem[{Felbo et~al.(2017)Felbo, Mislove, S{\o}gaard, Rahwan, and
  Lehmann}]{felbo2017}
Bjarke Felbo, Alan Mislove, Anders S{\o}gaard, Iyad Rahwan, and Sune Lehmann.
  2017.
\newblock Using millions of emoji occurrences to learn any-domain
  representations for detecting sentiment, emotion and sarcasm.
\newblock In \emph{Conference on Empirical Methods in Natural Language
  Processing (EMNLP)}.

\bibitem[{Fleiss(1971)}]{fleiss1971measuring}
Joseph~L Fleiss. 1971.
\newblock Measuring nominal scale agreement among many raters.
\newblock \emph{Psychological bulletin}, 76(5):378.

\bibitem[{Geng(2016)}]{geng2016label}
Xin Geng. 2016.
\newblock Label distribution learning.
\newblock \emph{IEEE Transactions on Knowledge and Data Engineering},
  28(7):1734--1748.

\bibitem[{Geng et~al.(2013)Geng, Yin, and Zhou}]{geng2013facial}
Xin Geng, Chao Yin, and Zhi-Hua Zhou. 2013.
\newblock Facial age estimation by learning from label distributions.
\newblock \emph{IEEE transactions on pattern analysis and machine
  intelligence}, 35(10):2401--2412.

\bibitem[{Henderson et~al.(2004)Henderson, Giese, and
  Cote}]{henderson2004impression}
Pamela~W Henderson, Joan~L Giese, and Joseph~A Cote. 2004.
\newblock Impression management using typeface design.
\newblock \emph{Journal of marketing}, 68(4):60--72.

\bibitem[{Juni and Gross(2008)}]{juni2008emotional}
Samuel Juni and Julie~S Gross. 2008.
\newblock Emotional and persuasive perception of fonts.
\newblock \emph{Perceptual and motor skills}, 106(1):35--42.

\bibitem[{Kar et~al.(2018)Kar, Maharjan, and Solorio}]{kar2018folksonomication}
Sudipta Kar, Suraj Maharjan, and Thamar Solorio. 2018.
\newblock Folksonomication: Predicting tags for movies from plot synopses using
  emotion flow encoded neural network.
\newblock In \emph{Proceedings of the 27th International Conference on
  Computational Linguistics}, pages 2879--2891.

\bibitem[{Kingma and Ba(2014)}]{kingma2014adam}
Diederik~P Kingma and Jimmy Ba. 2014.
\newblock Adam: A method for stochastic optimization.
\newblock \emph{arXiv preprint arXiv:1412.6980}.

\bibitem[{Kulahcioglu and De~Melo(2018)}]{kulahcioglu2018fontlex}
Tugba Kulahcioglu and Gerard De~Melo. 2018.
\newblock Fontlex: A typographical lexicon based on affective associations.
\newblock In \emph{Proceedings of the Eleventh International Conference on
  Language Resources and Evaluation (LREC-2018)}.

\bibitem[{Kullback and Leibler(1951)}]{kullback1951information}
Solomon Kullback and Richard~A Leibler. 1951.
\newblock On information and sufficiency.
\newblock \emph{The annals of mathematical statistics}, 22(1):79--86.

\bibitem[{Mackiewicz(2007)}]{mackiewicz2007audience}
Jo~Mackiewicz. 2007.
\newblock Audience perceptions of fonts in projected powerpoint text slides.
\newblock \emph{Technical communication}, 54(3):295--307.

\bibitem[{Mackiewicz and Moeller(2005)}]{mackiewicz2005people}
Jo~Mackiewicz and Rachel Moeller. 2005.
\newblock Why people perceive typefaces to have different personalities.
\newblock In \emph{International Professional Communication Conference, 2004.
  IPCC 2004. Proceedings.}, pages 304--313. IEEE.

\bibitem[{Mallinson et~al.(2017)Mallinson, Sennrich, and
  Lapata}]{mallinson2017paraphrasing}
Jonathan Mallinson, Rico Sennrich, and Mirella Lapata. 2017.
\newblock Paraphrasing revisited with neural machine translation.
\newblock In \emph{Proceedings of the 15th Conference of the European Chapter
  of the Association for Computational Linguistics: Volume 1, Long Papers},
  pages 881--893.

\bibitem[{Mohammad(2018{\natexlab{a}})}]{vad-acl2018}
Saif~M. Mohammad. 2018{\natexlab{a}}.
\newblock Obtaining reliable human ratings of valence, arousal, and dominance
  for 20,000 english words.
\newblock In \emph{Proceedings of The Annual Conference of the Association for
  Computational Linguistics (ACL)}, Melbourne, Australia.

\bibitem[{Mohammad(2018{\natexlab{b}})}]{LREC18-AIL}
Saif~M. Mohammad. 2018{\natexlab{b}}.
\newblock Word affect intensities.
\newblock In \emph{Proceedings of the 11th Edition of the Language Resources
  and Evaluation Conference (LREC-2018)}, Miyazaki, Japan.

\bibitem[{Mohammad and Turney(2013)}]{mohammad2013crowdsourcing}
Saif~M Mohammad and Peter~D Turney. 2013.
\newblock Crowdsourcing a word--emotion association lexicon.
\newblock \emph{Computational Intelligence}, 29(3):436--465.

\bibitem[{Nguyen et~al.(2016)Nguyen, Halpern, Wallace, and
  Lease}]{nguyen2016probabilistic}
An~Thanh Nguyen, Matthew Halpern, Byron~C Wallace, and Matthew Lease. 2016.
\newblock Probabilistic modeling for crowdsourcing partially-subjective
  ratings.
\newblock In \emph{Fourth AAAI Conference on Human Computation and
  Crowdsourcing}.

\bibitem[{O'Donovan et~al.(2014)O'Donovan, L{\=\i}beks, Agarwala, and
  Hertzmann}]{o2014exploratory}
Peter O'Donovan, J{\=a}nis L{\=\i}beks, Aseem Agarwala, and Aaron Hertzmann.
  2014.
\newblock Exploratory font selection using crowdsourced attributes.
\newblock \emph{ACM Transactions on Graphics (TOG)}, 33(4):92.

\bibitem[{Pennington et~al.(2014)Pennington, Socher, and
  Manning}]{pennington2014glove}
Jeffrey Pennington, Richard Socher, and Christopher Manning. 2014.
\newblock Glove: Global vectors for word representation.
\newblock In \emph{Proceedings of the 2014 conference on empirical methods in
  natural language processing (EMNLP)}, pages 1532--1543.

\bibitem[{Rodrigues et~al.(2014)Rodrigues, Pereira, and
  Ribeiro}]{rodrigues2014sequence}
Filipe Rodrigues, Francisco Pereira, and Bernardete Ribeiro. 2014.
\newblock Sequence labeling with multiple annotators.
\newblock \emph{Machine learning}, 95(2):165--181.

\bibitem[{Salminen et~al.(2018)Salminen, Al-Merekhi, Dey, and
  Jansen}]{salminen2018inter}
Joni~O Salminen, Hind~A Al-Merekhi, Partha Dey, and Bernard~J Jansen. 2018.
\newblock Inter-rater agreement for social computing studies.
\newblock In \emph{2018 Fifth International Conference on Social Networks
  Analysis, Management and Security (SNAMS)}, pages 80--87. IEEE.

\bibitem[{Shinahara et~al.(2019)Shinahara, Karamatsu, Harada, Yamaguchi, and
  Uchida}]{shinahara2019serif}
Yuto Shinahara, Takuro Karamatsu, Daisuke Harada, Kota Yamaguchi, and Seiichi
  Uchida. 2019.
\newblock Serif or sans: Visual font analytics on book covers and online
  advertisements.
\newblock \emph{arXiv preprint arXiv:1906.10269}.

\bibitem[{Shirani et~al.(2019)Shirani, Dernoncourt, Asente, Lipka, Kim,
  Echevarria, and Solorio}]{shirani2019learning}
Amirreza Shirani, Franck Dernoncourt, Paul Asente, Nedim Lipka, Seokhwan Kim,
  Jose Echevarria, and Thamar Solorio. 2019.
\newblock Learning emphasis selection for written text in visual media from
  crowd-sourced label distributions.
\newblock In \emph{Proceedings of the 57th Annual Meeting of the Association
  for Computational Linguistics}, pages 1167--1172.

\bibitem[{Srinivasan and Chander(2019)}]{srinivasan2019crowdsourcing}
Ramya Srinivasan and Ajay Chander. 2019.
\newblock Crowdsourcing in the absence of ground truth--a case study.
\newblock \emph{arXiv preprint arXiv:1906.07254}.

\bibitem[{Urkullu et~al.(2019)Urkullu, Perez, and
  Calvo}]{urkullu2019evaluation}
A~Urkullu, A~Perez, and B~Calvo. 2019.
\newblock On the evaluation and selection of classifier learning algorithms
  with crowdsourced data.
\newblock \emph{Applied Soft Computing}, 80:832--844.

\bibitem[{Wolf et~al.(2019)Wolf, Debut, Sanh, Chaumond, Delangue, Moi, Cistac,
  Rault, Louf, Funtowicz, and Brew}]{Wolf2019HuggingFacesTS}
Thomas Wolf, Lysandre Debut, Victor Sanh, Julien Chaumond, Clement Delangue,
  Anthony Moi, Pierric Cistac, Tim Rault, R'emi Louf, Morgan Funtowicz, and
  Jamie Brew. 2019.
\newblock Huggingface's transformers: State-of-the-art natural language
  processing.
\newblock \emph{ArXiv}, abs/1910.03771.

\bibitem[{Yang et~al.(2018)Yang, Drake, Damianou, and
  Maarek}]{yang2018leveraging}
Jie Yang, Thomas Drake, Andreas Damianou, and Yoelle Maarek. 2018.
\newblock Leveraging crowdsourcing data for deep active learning an
  application: Learning intents in alexa.
\newblock In \emph{Proceedings of the 2018 World Wide Web Conference}, pages
  23--32. International World Wide Web Conferences Steering Committee.

\bibitem[{Yang et~al.(2015)Yang, Gao, Xing, Huo, Wei, Zhou, Wu, and
  Geng}]{yang2015deep}
Xu~Yang, Bin-Bin Gao, Chao Xing, Zeng-Wei Huo, Xiu-Shen Wei, Ying Zhou, Jianxin
  Wu, and Xin Geng. 2015.
\newblock Deep label distribution learning for apparent age estimation.
\newblock In \emph{Proceedings of the IEEE international conference on computer
  vision workshops}, pages 102--108.

\end{thebibliography}
\bibliographystyle{acl_natbib}
\end{document}